%% file: MAIN.tex
\documentclass[sigconf]{acmart} 

\usepackage{booktabs} 

\usepackage{hyperref}
\usepackage{amsmath,amssymb,mathtools}
\usepackage{xr}

\copyrightyear{2018} 
\acmYear{2018} 
\setcopyright{acmcopyright}
\acmConference[GECCO '18]{Genetic and Evolutionary Computation Conference}{July 15--19, 2018}{Kyoto, Japan}
\acmBooktitle{GECCO '18: Genetic and Evolutionary Computation Conference, July 15--19, 2018, Kyoto, Japan}
\acmPrice{15.00}
\acmDOI{10.1145/3205455.3205529}
\acmISBN{978-1-4503-5618-3/18/07}

\begin{document}

\title{Interoceptive robustness through environment-mediated morphological development}

\author{Sam Kriegman}
\affiliation{%
  \institution{University of Vermont}
  \city{Burlington} 
  \state{VT} 
  \country{USA}
}
\email{sam.kriegman@uvm.edu}

\author{Nick Cheney}
\affiliation{%
  \institution{University of Wyoming}
  \city{Laramie} 
  \state{WY} 
  \country{USA}
}

\author{Francesco Corucci}
\affiliation{%
  \institution{3DNextech s.r.l.}
  \city{Livorno} 
  \country{Italy}
}

\author{Josh C. Bongard}
\affiliation{%
  \institution{University of Vermont}
  \city{Burlington} 
  \state{VT} 
  \country{USA}
}

\renewcommand{\shortauthors}{Kriegman et al.}

\begin{teaserfigure}
\centering
\vspace{-1.75em}
\includegraphics[width=\linewidth]{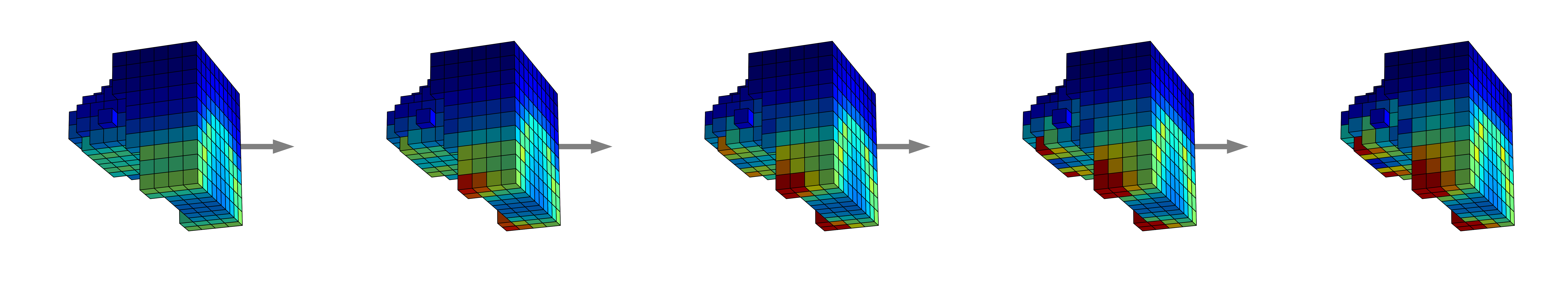} \\
\vspace{-1.75em}
\includegraphics[width=0.35\linewidth]{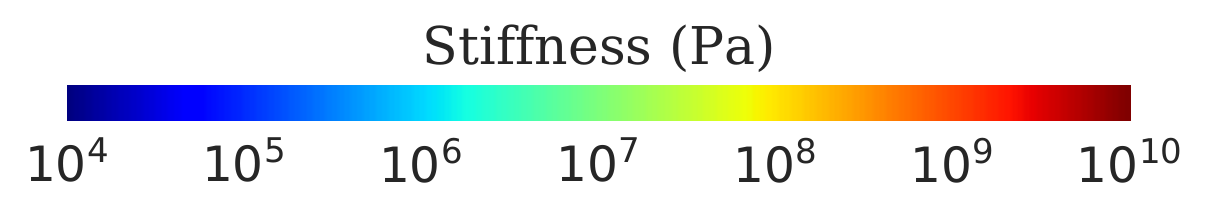} \\
\vspace{-1em}
\caption{A single robot grows calluses as it walks, 
in response to pressure on its feet   
(\href{https://youtu.be/0cmwpcxSUWI}{\textcolor{blue}{\textbf{\texttt{youtu.be/0cmwpcxSUWI}}}}).
} 
\label{fig:calluses}
\vspace{1em}
\end{teaserfigure}

\input{abstract}

\maketitle

\input{intro}

\input{methods}

\input{results}

\input{discussion}

\input{acknowledgments}

\bibliographystyle{ACM-Reference-Format}
\bibliography{main}

\end{document}

%% file: abstract.tex
\begin{abstract}

Typically, AI researchers and roboticists try to realize
intelligent behavior in machines by tuning parameters of a 
predefined structure (body plan and/or neural network
architecture) using evolutionary or learning algorithms. 
Another but not unrelated longstanding property of these systems is their brittleness to slight aberrations, as highlighted by the growing deep learning literature on adversarial examples.
Here we show robustness can be achieved by
evolving the 
geometry of soft robots, their
control systems, and how
their material properties develop
in response to one particular interoceptive stimulus
(engineering stress) during their lifetimes.
By doing so we realized robots that 
were equally fit but more robust to 
extreme material defects (such as 
might occur during fabrication or by damage thereafter)
than robots that did not develop during their lifetimes,
or developed in response to a different interoceptive
stimulus (pressure).
This suggests that the interplay between changes
in the containing systems
of agents (body plan and/or neural architecture)
at different temporal scales (evolutionary
and developmental) along different modalities
(geometry, material properties, synaptic weights)
and in response to different signals (interoceptive
and external perception) all
dictate those agents' abilities to evolve or 
learn capable and robust strategies.
\end{abstract}

%
%

\ccsdesc[500]{Computing methodologies~Mobile agents}

\keywords{Soft robotics.}

%% file: intro.tex
\section{Introduction}
\label{sec:introduction}

A major characteristic of life is that three broad time scales are relevant to it: evolution, development and physiological functioning.
Engineered systems, in marked contrast, often employ an evolutionary or learning algorithm to improve their behavior over time, but rarely employ morphological development; any changes to the physical layout are made in between evaluations \citep{sims1994evolving,lipson2000automatic,cheney2013unshackling}, if they are made at all.

Two notable exceptions are modular robots \citep{zykov2005robotics}, which may reconfigure their bodies by adding and removing discrete structures, and soft robots \citep{shepherd2011multigait}, which may continuously alter the local volumes of different parts of their
bodies while behaving. 
Others \citep{bongard2011morphological} have approximated topological change in rigid bodies by extending outward and angling downward 
appendages using a combination of linear and rotary actuators, thus simulating limb growth.

Several computational but embodied models of \textit{prenatal} development have been reported in the literature \cite{dellaert1996developmental,
Eggenberger97,
Bongard01,
miller2004evolving,
doursat2009organically}.
As implied,
cellular growth therein occurred prior to any physiological functioning. 
Thus, these studies included change during only
two of the three time scales relevant to life:
evolutionary and behavioral change, but not postnatal developmental
change.

The most common argument in favor of development is that some aspects of the environment are unpredictable, so it is advantageous to leave some decisions up to development rather than specifying them genetically.
Although self evident, it remains to determine which mechanisms of development should be instantiated in robots to realize plastic, adaptive, and useful machines.

Naturally, the performance of an evolved system depends on its capacity for evolutionary improvement: its evolvability.
Development can, under certain conditions, smooth the search space evolution that operates in, thus increasing evolvability.
This process, known as the Baldwin effect \citep{baldwin1896new,dennett2003baldwin}, starts with an advantageous characteristic acquired during the development of individuals, such as the callouses in Fig. \ref{fig:calluses}. 
This can create a new gradient in the evolutionary search space, rewarding descendants that more rapidly manifest the trait during their lifetimes \citep{hinton1987learning,kriegman2017morphological}
and retain it through the remainder of their lifetime \citep{kriegman2017minimal}.
Assuming such mutations exist and can be naturally selected \citep{kriegman2017morphological}, following the gradient requires incrementally reducing development in the manifold of the search space that can express variations on the trait \cite{waddington1942canalization}.

However, fitness landscapes that evolution climbs, and development sometimes smooths, tend not to remain static in realistic settings.
On this vacillating landscape, when the best thing to do does not remain the same, a highly evolvable but non-robust system will need to keep starting over from scratch every time the conditions change.
Computational and engineered systems provide countless examples of systems with nearly perfect performance in a controlled environment, such as a factory, but who turn out to be (often comically) brittle to slight changes in their internal structure, such as damage, or their external environment such as moving on to new terrain or transferal from simulation to reality 
\citep{carlson2005ugvs,
koos2013transferability,
goodfellow2013empirical,
szegedy2013intriguing,
athalye2017synthesizing,
nguyen2015deep}.

Although generally absent from engineered systems (but see \cite{bongard2006resilient,cully2015robots}), the canonical form of robustness is seen to some extent in all organisms, and it comes from the act of development itself.
For example, a plant that grows according to a fixed program will capture less light than a plant that grows toward sunlight.
But there is another, more subtle form of robustness that we will refer to as `intrinsic robustness' because it is a property of a system's structure rather than of the process by which it may change.

Developmental change produces intrinsically robust systems because they evolved from designs that had to maintain adequate performance along additional dimensions of change \citep{bongard2011morphological,kriegman2017morphological}.
Through morphological development specifically, evolution is compelled to maintain designs that are capable across a series of body plans, with different 
sensor-motor contingencies; and the ability to tolerate such perturbations can become inherited to some extent in descendants' behaviors \cite{bongard2011morphological} and morphologies \cite{kriegman2017morphological}, even when their developmental flexibility is reduced or completely removed by canalization or fabrication.

And yet, despite the ubiquity of
morphological development in nature, and the adaptive advantages it seemingly confers, there are only a handful of cases reported in the literature in which a simulated robot's mechanical structure was allowed to change while it was behaving 
(e.g.~\citep{ventrella1998designing,
komosinski2003framsticks,
bongard2011morphological,
kriegman2017morphological,
kriegman2017minimal}),
all of which modeled morphological development as a genetically predetermined process: the environment could not influence the way in which development unfolded.

Assuming that an engineered system is capable of local morphological change in response to environmental signals, it is unclear how it should do so, beyond the examples of morphological plasticity observed in nature. 
Examples include Wolff's law \citep{ruff2006s}---bone grows in response to particular mechanical loading profiles---and Davis' law---soft tissue increases in strength in response to intermittent mechanical demands.
One can envisage other such laws that are not known to occur in biology but could be helpful in a specific artificial system, such as end effectors softening in response to pressure, which might enhance their ability to safely manipulate irregular or delicate objects \cite{brown2010universal}.
Indeed, the genesis of the work presented here is one such anecdotal example given in \citep{corucci2017evolutionary}, where a single robot, subjected to an abrupt doubling in gravity, stiffened its body in reaction to the increased pressure.
However, whether or how it could provide a behavioral advantage, nor whether pressure is the best interoceptive signal to developmentally respond to,
was not investigated.

As a step towards a more adequate picture, we introduce here a simple form of a developmental feedback mechanism:
Genetic systems dictate how organisms develop in response to interoceptive stimuli, and development alters the kinds of interoceptive conditions the organism experiences.
More specifically, at every time step, the proposed model of closed-loop development:
\begin{enumerate}
\item `listens to' load signatures generated from movement; and, in response,
\item modifies the robot's rigidity,
\end{enumerate}
which will change the way it distributes load and generates movement at the next time step.

Optimizing a system that may form a continuum of rigid and soft components---and in which this admixture may change over time---is extremely nonintuitive and underexplored.
Thus, a study of the adaptive properties of such systems---and how they can best be optimized to render useful work---is initiated here.

%% file: methods.tex
\section{Methods}
\label{sec:methods}

We evolved locomotive machines constructed from voxels with heterogeneous stiffness. 
Like many organisms \citep{ruff2006s}, the robot's material stiffness progressively changes in response to mechanical loading incurred as the robot behaves.
This ontogenetic change occurs independently at each voxel according to an evolved local rule.

\subsection*{Physical simulation.}

The soft-matter physics engine \textit{Voxelyze} \citep{hiller2014dynamic} is used to calculate the movement of robots resulting from their interaction with a virtual 3D terrestrial environment.
Each robot is simulated for 25 times the length of an expansion/contraction cycle (a total of five seconds). 
The displacement between the starting coordinates and the agent's final center of mass (in the $xy$ plane) is recorded.\footnote{\href{https://github.com/skriegman/2018-gecco}{\textcolor{blue}{\textbf{\texttt{github.com/skriegman/2018-gecco}}}} contains the source code necessary for reproducing the results reported in this paper.}


\subsection*{Heavy materials.}


Materials are simulated to have increased mass relative to those used by \cite{hiller2012automatic,cheney2013unshackling,cheney2014electro}.
Constructed from heavier materials, many previously mobile robots become crushed under their own weight and require stiffer material to support locomotion with the same geometry.
However, we also restricted actuation amplitude as materials grow stiffer to better approximate the properties of real materials with different stiffnesses.
This creates an interesting and realistic trade-off: the stiffest material can easily support any body plan but cannot move on its own (like a skeleton without muscle), whereas the softest material can readily elicit forward movement in smaller bodies but cannot support many larger and potentially faster-moving body plans, 
such as those with narrow supporting limbs.
Thus, a robot must carefully balance support with actuation.

\subsection*{Quad-CPPN encoding: $\mathbf{\mathbb{C}_1,\mathbb{C}_2,\mathbb{C}_3,\mathbb{C}_4}$
}

Following \cite{cheney2013unshackling}, robot physiology is genetically encoded by a Compositional Pattern Producing Network (CPPN) \citep{stanley2007cppn}, a scale-free mapping that biases search toward symmetrical and regular patterns which are known to facilitate locomotion.

Each point on a 10$\times$10$\times$10 lattice is queried by its cartesian coordinates in 3D space and its radial distance from the lattice center.
An evolved CPPN takes these coordinates as input and returns a single value which is used to set some property of that point in the workspace.
We used four independent CPPNs to separately encode: Geometry, Stiffness, Development, and Actuation.

\subsection*{$\mathbf{\mathbb{C}_1}\;$ Geometry.}
The geometry of a robot is specified by a bitstring that indicates whether material is present (1) or absent (0) at each lattice point in the workspace, as dictated by $\mathbb{C}_1$.
The robot's geometric shape is taken to be the largest contiguous collection of present voxels.

\subsection*{$\mathbf{\mathbb{C}_2}\;$ Stiffness.}
Young's modulus is often used as an approximate measure of material stiffness.
Robots are typically constructed of materials such as metals and hard plastics that have moduli in the order of $10^9 - 10^{12}$ pascals (Pa), whereas soft robots (and natural organisms) are often composed of materials with moduli in the order of
$10^4 - 10^9$ Pa \citep{rus2015design}.

Here, voxels may have moduli in the range $10^4 - 10^{10}$ Pa.
The robot's congenital stiffness is set at each voxel by $\mathbb{C}_2$, but may be changed by development.


\subsection*{$\mathbf{\mathbb{C}_3}\;$ Development.}
The robot's stiffness distribution $k$ 
can change progressively during its lifetime $t$ in response to localized engineering stress $\sigma$ or pressure $p$.
The rate of change $\alpha_i$ is specified at the $i^{\text{th}}$ voxel by $\mathbb{C}_3$, with possible values in $0\pm10$.
We compare three developmental variants.
\begin{align}
&\textit{\textbf{None:}} 
\hspace{2em} &\frac{dk_i}{dt}& \,=\, 0 
\hspace{10em}
\label{eq:none} \\[0.35em]
&\textit{\textbf{Stress:}} \hspace{3em} &\frac{dk_i}{dt}& \,=\, \alpha_i \cdot \frac{d\sigma_i}{dt} 
\label{eq:stress} \\[0.35em]
&\textit{\textbf{Pressure:}} \hspace{3em} &\frac{dk_i}{dt}& \,=\, \alpha_i \cdot \frac{dp_i}{dt} 
\label{eq:pressure}
\end{align}

\subsection*{$\mathbf{\mathbb{C}_4}\;$ Actuation.}
Robots are `controlled by' volumetric actuation: a sinusoidal expansion/contraction of each voxel with a maximum amplitude of 50\% volumetric change.\footnote{The scare quotes are intended to highlight the fact that actuation policies no more dictate the movement of a robot than its geometry.}
However, linear damping $\xi$ is implemented into the system such that the stiffest material does not actuate (Eq. \ref{eq:damp}). 
The phase difference $\phi_i$ of each voxel is determined by $\mathbb{C}_4$, which offsets its oscillation relative to a central pattern generator.

Prior to actuation, each voxel has a resting length of one centimeter.
This length is periodically varying ($f=$ 5 Hz) by approximately 14.5\%
($A\approx$ 0.145 cm), but damped by $\xi$.
The instantaneous length of the $i$\textsuperscript{th} voxel is thus:
\begin{equation}
\label{eq:actuation}
\psi_i(t) = 1+A \cdot \sin(2\pi f t + \phi_i) \cdot \xi(k_i) \, ,
\end{equation}
where:
\begin{equation}
\label{eq:damp}
\xi(k_i) = \frac{k_{\max} - k_i}{k_{\max}-k_{\min}} \, .
\end{equation}

\subsection*{Evolution.}

We employed a standard evolutionary algorithm, Age-Fitness-Pareto Optimization \citep{Schmidt2011}, which uses the concept of Pareto dominance and an objective of age (in addition to fitness) intended to promote diversity among candidate designs. 

We performed twenty independent evolutionary trials with different random seeds (1-20); in each trial, a population of 24 robots was evolved for five thousand generations.
Every generation, the population is first doubled by creating modified copies of each individual in the population. 
The age of each individual is then incremented by one.
Next, an additional random individual (with age zero) is injected into the population (which now consists of 49 robots). 
Finally, selection reduces the population down to its original size (24 robots) according to the two objectives of fitness (maximized) and age (minimized).

Mutations add/remove/alter a particular node/link of a CPPN.
They are applied by first selecting which networks to mutate, with the possibility to select all four, and then choosing which operations to apply to each.


\subsection*{Hypothesis testing.}

We use bootstrapping to construct hypothesis tests.
All $P$-values are reported with Bonferroni correction for multiple (typically three) comparisons.
We adopt the following convention: `${\ast}{\ast}{\ast}$' for ${P<0.001;}$ `${\ast}{\ast}$' for $P<0.01;$ and `${\ast}$' for $P<0.05$.

%% file: results.tex
\externaldocument{methods}
\externaldocument{discussion}

\section{Results}
\label{sec:results}

Videos of all sixty run champions (pictured in Figs. \ref{fig:none}, \ref{fig:stress}, \ref{fig:pressure}) can be seen at
\href{https://www.youtube.com/playlist?list=PL7qssg0uLKTaFhaCRC0WviaGeOitOE8MR}{\textcolor{blue}{\textbf{\texttt{goo.gl/T5wZNQ}}}}.

\subsection*{Evolvability.}

\begin{figure}
\centering
{\Large \textbf{Non-developmental champions (Eq. \ref{eq:none})}}
\includegraphics[width=\linewidth]{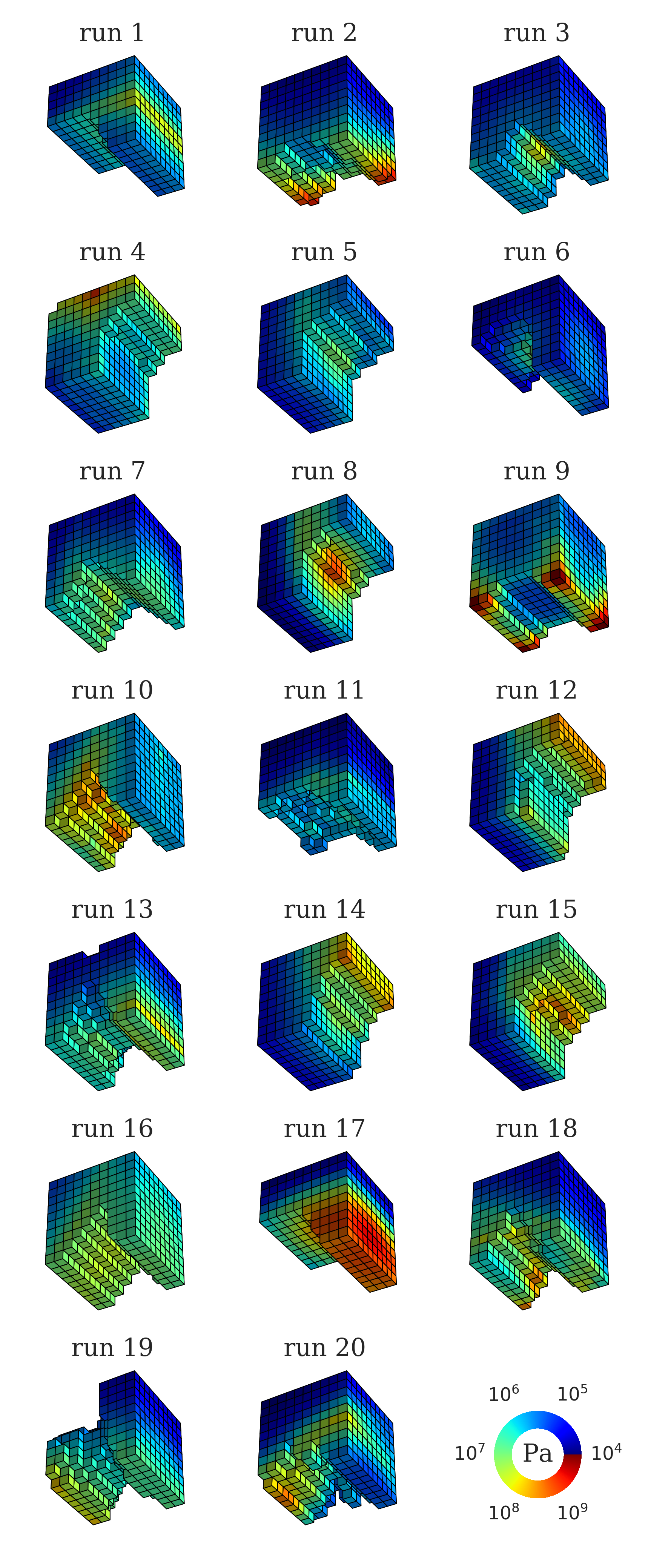}
\vspace{-2.5em}
\caption{\label{fig:none} Run champions colored by congenital stiffness, which ranges from 10\textsuperscript{4} to 10\textsuperscript{10} Pa. After settling under gravity, robots move toward the right-hand side of the page.}
\end{figure}

\begin{figure}
\centering
{\Large \textbf{Stress-adaptive champions (Eq. \ref{eq:stress})}}
\includegraphics[width=\linewidth]{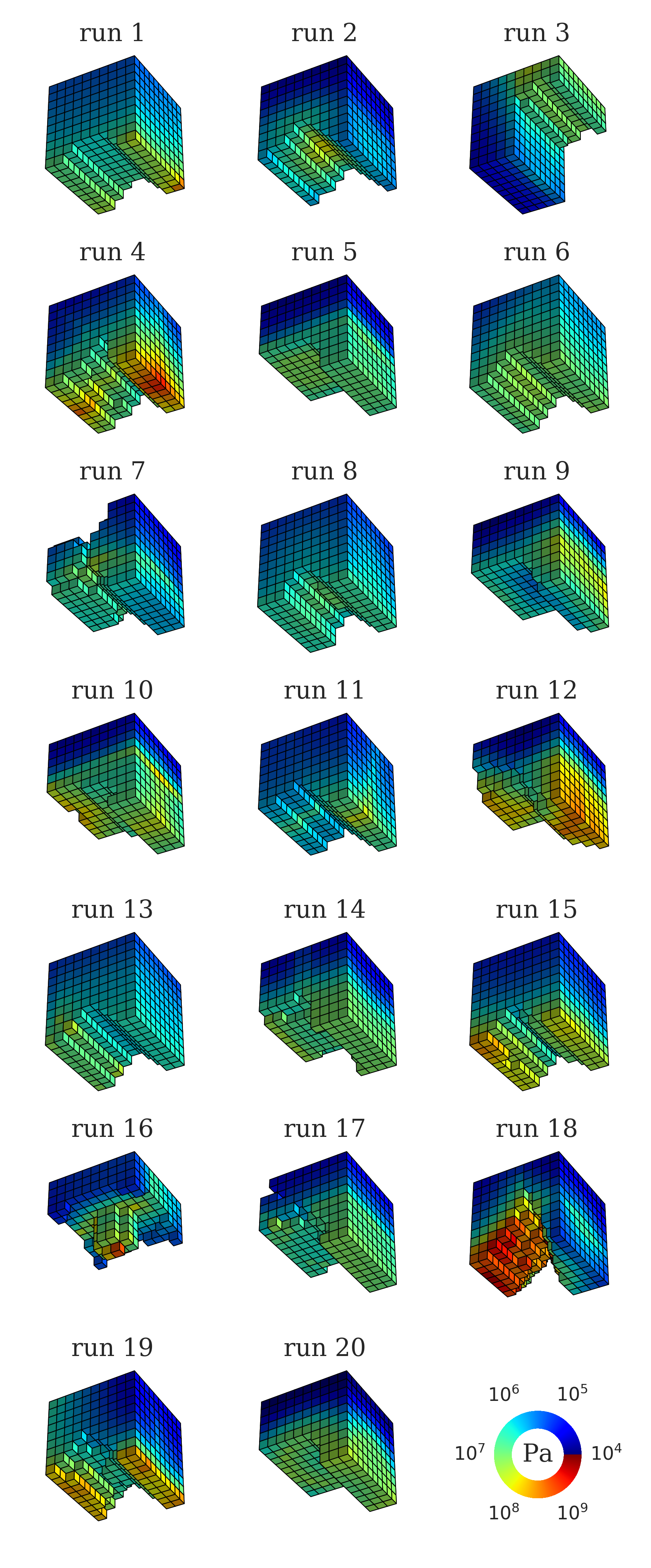}
\vspace{-2.5em}
\caption{\label{fig:stress} Run champions colored by congenital stiffness which can change during operation (ontogeny) in response to \textit{engineering stress.}}
\end{figure}

\begin{figure}
\centering
{\Large \textbf{Pressure-adaptive champions (Eq. \ref{eq:pressure})}}
\includegraphics[width=\linewidth]{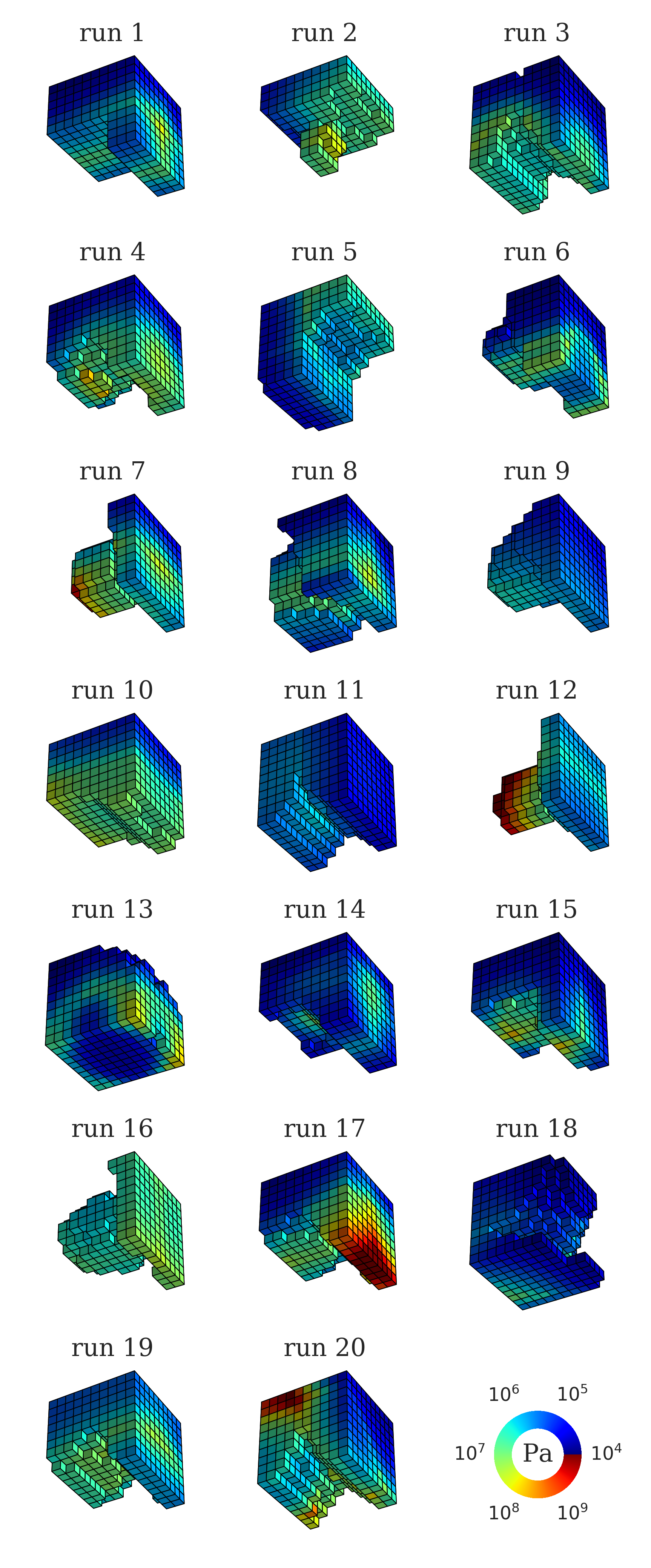}
\vspace{-2.5em}
\caption{\label{fig:pressure} Run champions are colored by congenital stiffness which can change during operation (ontogeny) in response to \textit{pressure.}}
\end{figure}

\begin{figure*}[t]
\centering 
\includegraphics[width=0.32\linewidth]{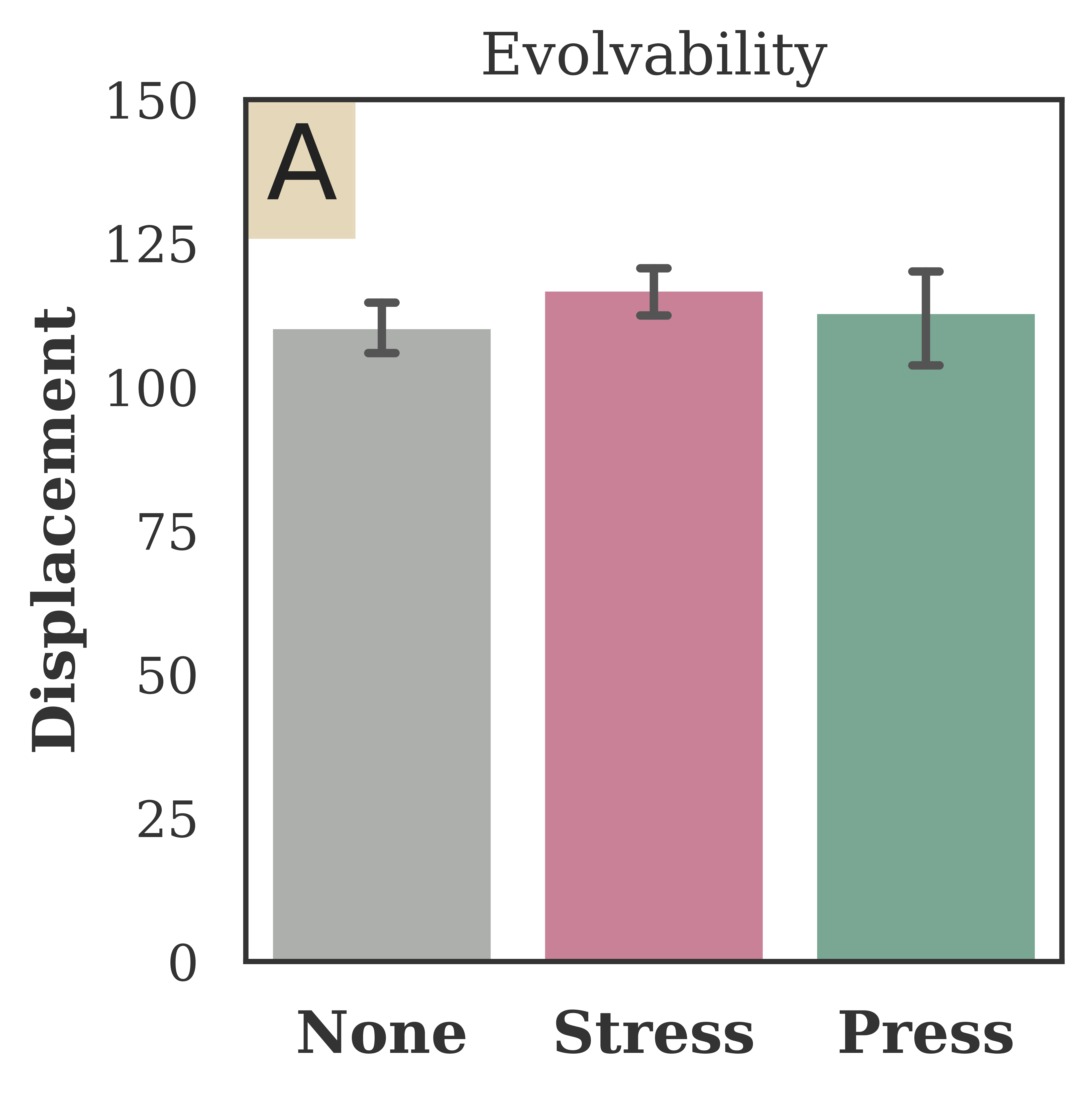}
\includegraphics[width=0.32\linewidth]{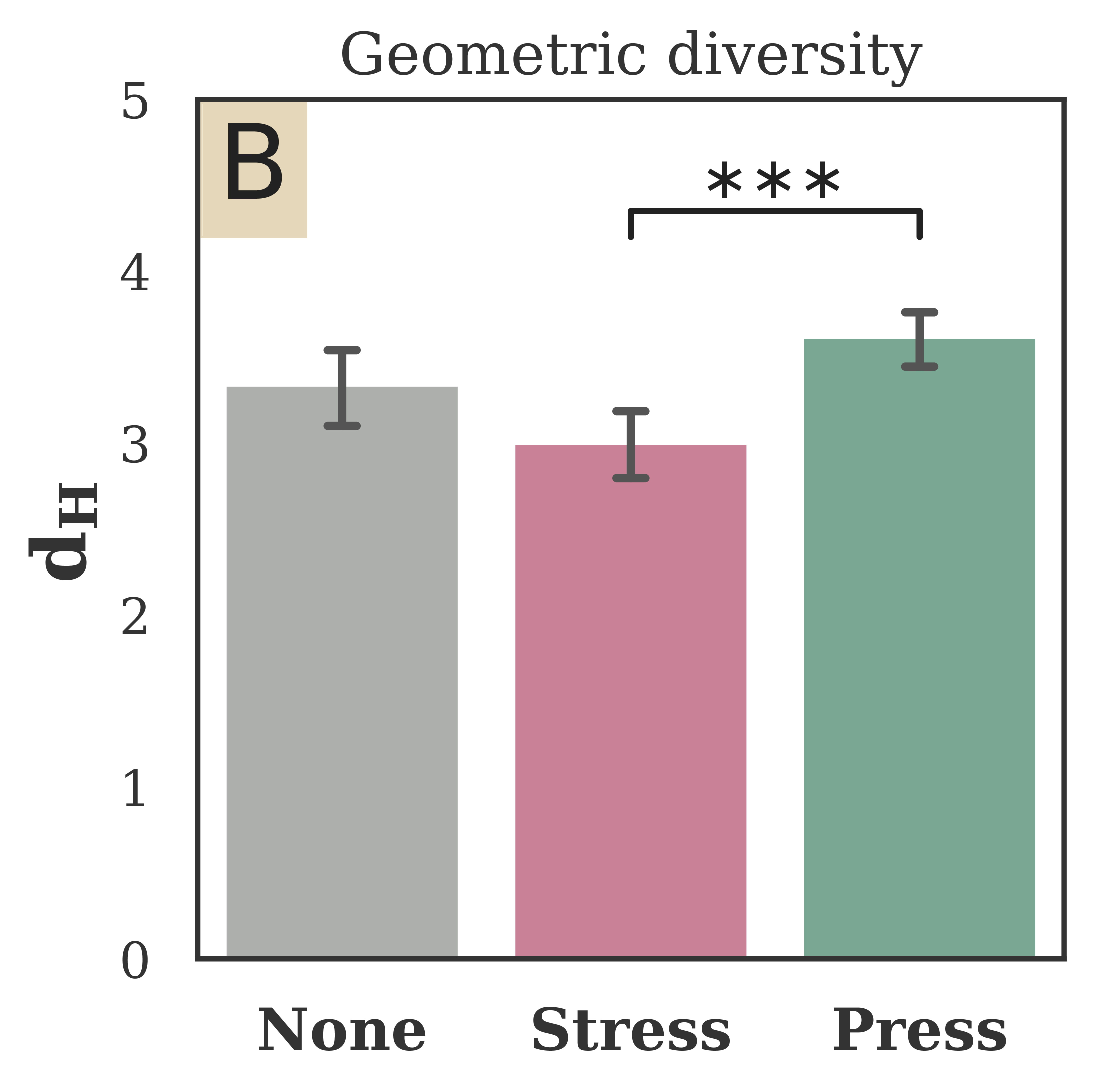} 
\includegraphics[width=0.32\linewidth]{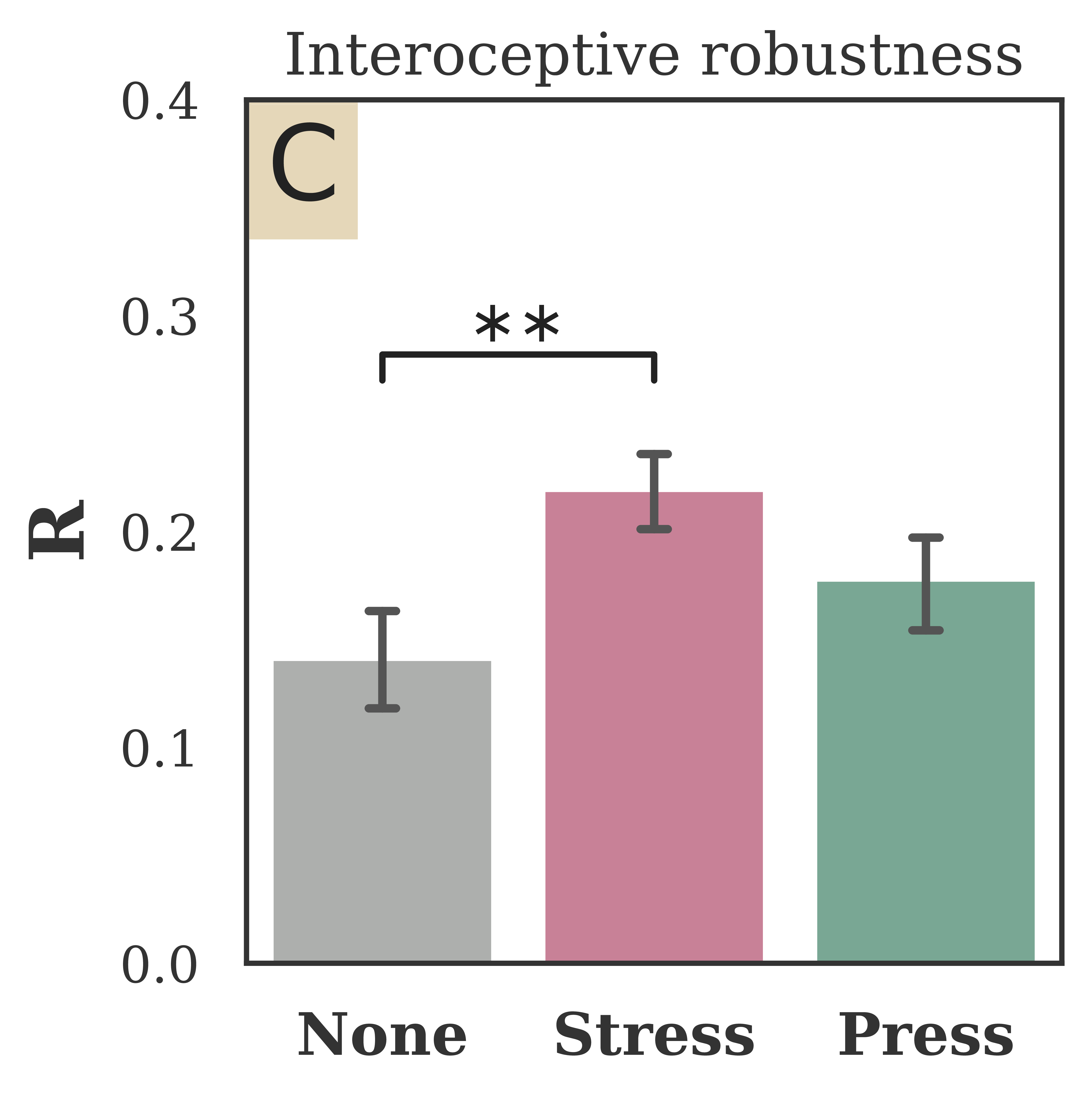}
\includegraphics[width=0.32\linewidth]{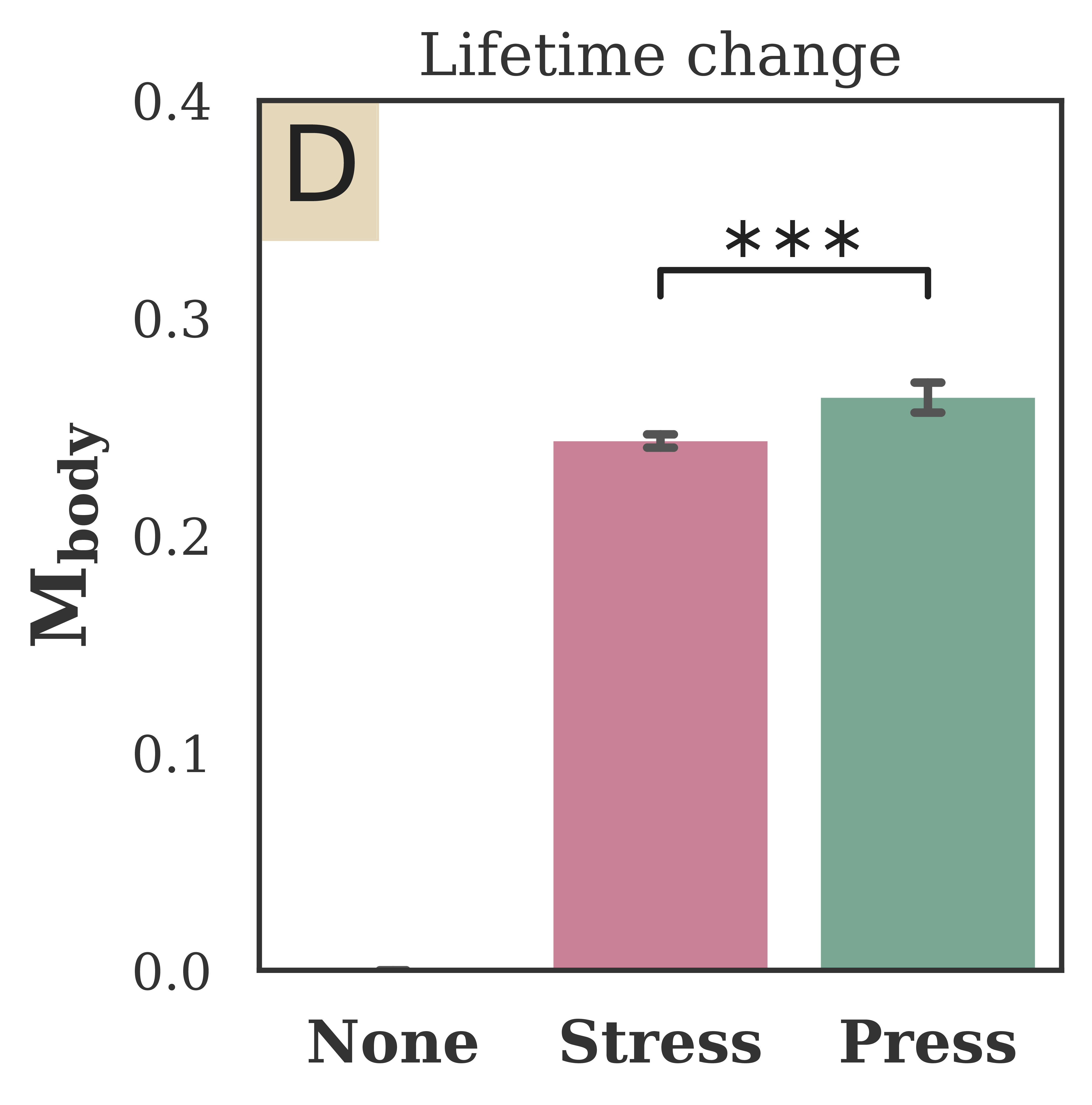}
\includegraphics[width=0.32\linewidth]{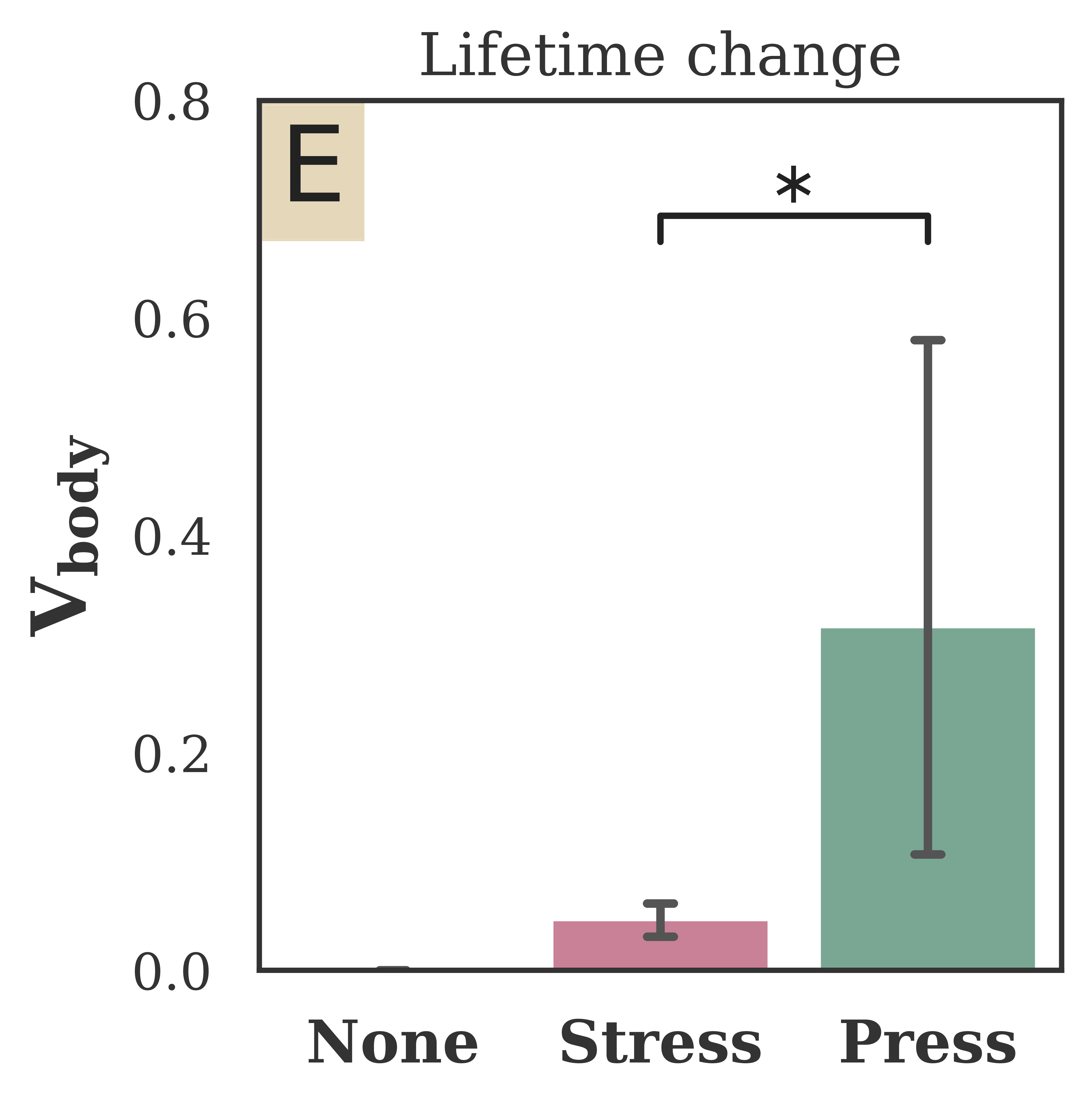}
\includegraphics[width=0.32\linewidth]{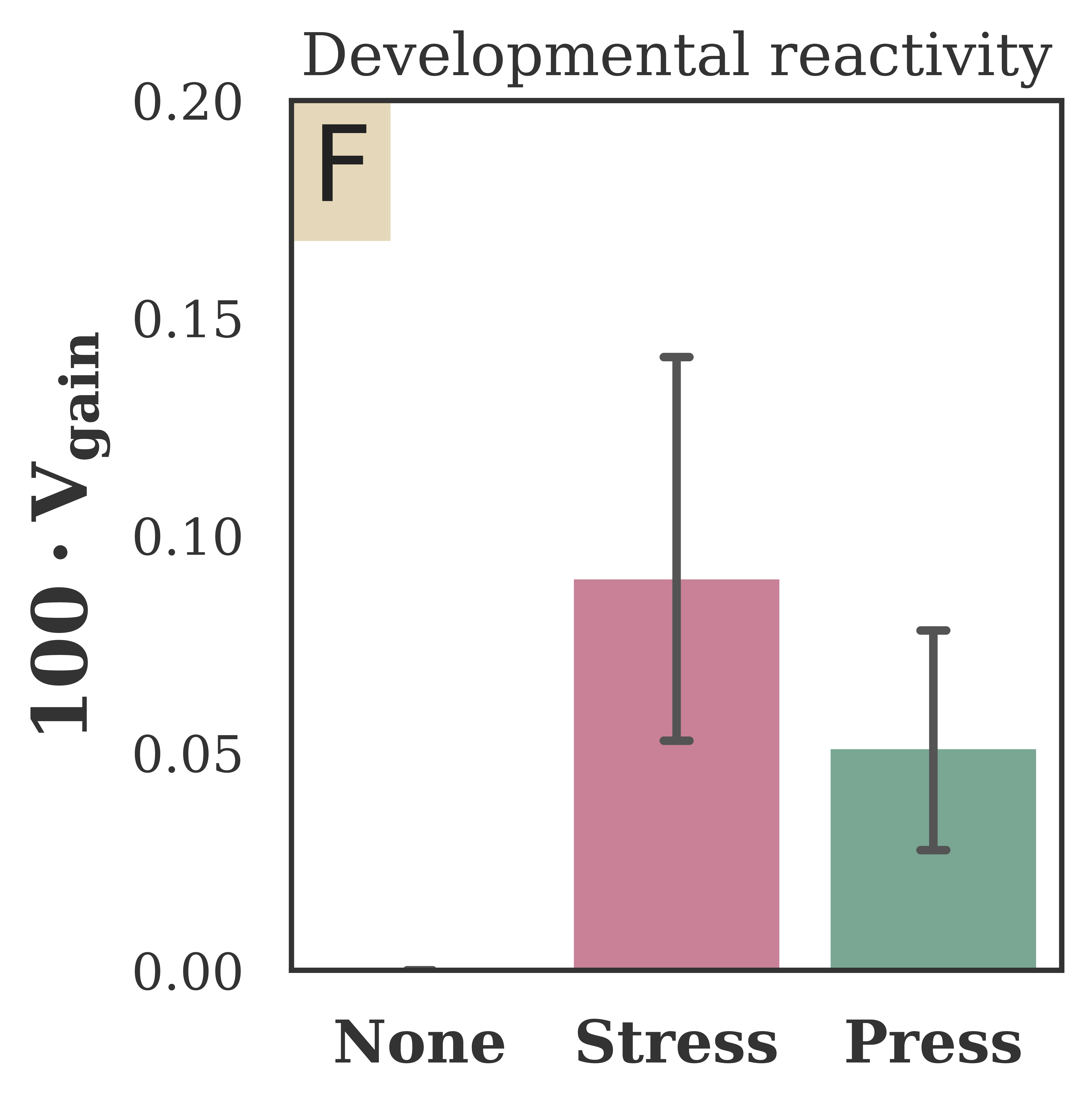}
\caption{\label{fig:run_champs} 
Means (with 95\% C.I.) for various statistics of the run champions, at generation 5000: 
(A) Fitness as the final displacement of a robot, measured by voxel-length units; 
(B) Diversity as the pairwise Hausdorff distances of robot geometries; 
(C) Robustness as the relative fitness (testing fitness divided by training fitness) after development is removed and a random stiffness distribution is introduced
into the champion's body (Eq. \ref{eq:robustness});
(D) Mean, taken across the body, of relative lifetime change in stiffness, as a measure of the lack of canalization (Eq. \ref{eq:mu}; lower bars indicate more canalization);
(E) Variance, taken across the body, of relative lifetime change in stiffness, as a measure of heterogeneity/nonuniformity in developmental reactions (Eq. \ref{eq:sigma});
(F) Variance, taken across the body, of the coefficients/gain of developmental reactivity (Eq. \ref{eq:sigma-gain}).
}
\end{figure*}

We first investigated whether environment-mediated morphological development affected evolvability (Fig. \ref{fig:run_champs}A).
At the termination of an evolutionary trial, we only consider the most fit individual{\textemdash}the run champion{\textemdash}from each of the twenty independent trials (Figs. \ref{fig:none}, \ref{fig:stress}, \ref{fig:pressure}).
After correcting for three comparisons, there was not enough evidence to reject the null hypothesis{\textemdash}that there is no difference between adaptive and nonadaptive robots, in terms of mean displacement{\textemdash}at the 0.05 level.

These results could be taken to suggest one of the following. 
Either the search space is sufficiently smooth prior to development (actuation and support are not as antagonistic as envisaged), or the proposed developmental mechanism is an insufficient smoothing 
mechanism
(successful ways to change stiffness as a linear function of stimuli are sparse in the search space).

\subsection*{Geometric diversity.}

We investigate morphological diversity next by
employing the Hausdorff distance $d_H$ as a metric to compare the similarity of two robot geometries, $A$ and $B$.
For each voxel in $A$, the closest voxel in $B$ is identified, according to euclidean distance $d$.
Similarly, for each voxel in $B$, the closest voxel in $A$ is identified.
The Hausdorff metric is the larger of these two distances. 
Formally,
\begin{equation}
\label{eq:hausdorff}
d_H(A,B) = \max\{\,\sup_{a \in A} \inf_{b \in B} d(a,b),\; \sup_{b \in B} \inf_{a \in A} d(a,b)\,\} \, . 
\end{equation}
Informally, two robots are close in the Hausdorff distance if every voxel of either robot is close to some voxel of the other robot.

We calculated the Hausdorff distance between each of the $\binom{20}{2}=190$ possible pairings of the 20 run champions (Fig. \ref{fig:run_champs}B).
Because $d_H(A,B)$ depends on the orientations of $A$ and $B$, we rotate $B$ in the $xy$ plane (0, 90, 180, and 270 degrees) and the $yz$ plane (0 and 90 degrees), and select the rotation that creates the smallest $d_H(A,B)$.

We found the evolved body shapes of pressure-adaptive robots to be more diverse than those of stress-adaptive robots $(P<0.001)$.
We did not find a significant difference, at the 0.05 level, between adaptive and nonadaptive treatments using this particular measure of morphological diversity.

Across all three treatments,
there appear on visual inspection to be three types of geometries (Figs. \ref{fig:none}-\ref{fig:pressure}): a $\Pi$ robot with wide posterior and anterior legs; a $\Gamma$ robot whose legs meet perpendicularly; and a $\Upsilon$ robot that connects a (mainly cylindrical) leg perpendicularly to the center of a $10\times10$ vertical plane.
Depending on how one counts, the $\Upsilon$ species can be seen in at most one nonadaptive robot (Fig. \ref{fig:none}, run 19), two stress-adaptive robots (Fig. \ref{fig:stress}, runs 7 and 16), and six pressure-adaptive robots (Fig. \ref{fig:pressure}, runs 2, 6, 7, 9, 12, and 16).
Pressure-adaptive robots have more diversity by virtue of more $\Upsilon$ robots.

\subsection*{Interoceptive robustness.}

To investigate the relative robustness (if any) across the three treatments,
in the following experiment, development was manually removed from the stress- and pressure-adaptive run champions.
We then tested the sensitivity of the resulting reduced robots to their evolved congenital stiffness distribution (Fig. \ref{fig:run_champs}C).
To do so, we replaced the evolved network dictating material stiffness, $\mathbb{C}_2,$ with a random number generator that draws from the same range of possible stiffness ($10^4 - 10^{10}$ Pa).
That is to say, we `built' the evolved run champions without any errors in the specifications of geometry and actuation, but completely ignored the evolved specifications of their material stiffness, replacing them instead with random noise.
We then calculated the relative fitness
\begin{equation}
\label{eq:robustness}
R = F_{\text{test}}\,/\,F_{\text{train}} \, ,
\end{equation}
where $F_{\text{train}}$ is the fitness achieved using the evolved stiffness and $F_{\text{test}}$ is the fitness when tested with a random stiffness distribution.
We repeated this process ten times for each run champion, each time drawing a new random stiffness distribution.

We found that, compared to nonadaptive robots, reduced stress-adaptive robots 
were 
more robust to this (extreme) discrepancy between training and testing stiffness distributions $(P < 0.01)$.
These results are consistent with the 
found correlation between development and robustness
\cite{miller2004evolving,bongard2011morphological,kriegman2017morphological}.
However, the results here indicate that this correlation is contingent on the kind of environmental
signal the developing agent responds to: there was no difference between pressure-adaptive and nonadaptive robots in this regard, at the 0.05 level.

This implies that by behaving interoceptively with respect to engineering stress, robots evolved the ability to ameliorate large deviations from their expected material properties, but by behaving interoceptively with respect to pressure, robots did not evolve this character.
Because development was manually removed beforehand, robustness in our case was not a matter of changing one's body, as in the example of plant growth \citep{sultan2000phenotypic}; rather, it is an intrinsic property of structure (geometries and actuation patterns) educed from ancestors who changed in response to one particular internal state (stress), but not from those who responded to another (pressure).

The difference in robustness between nonadaptive robots and stress-adaptive, but not pressure-adaptive robots, could be due in part to the fact that there are simply more pressure-adaptive $\Upsilon$ robots than stress-adaptive $\Upsilon$ robots.
While $\Upsilon$ robots tend to be more fit than $\Pi$ and $\Gamma$ robots $(P_{\Pi}<0.05;\, P_{\Gamma}<0.05)$,
they also appear to exploit their material properties to a greater degree, and are thus more sensitive to changes in its constitution, compared to $\Pi$ and $\Gamma$ robots $(P_{\Pi}<0.05;\, P_{\Gamma}<0.01)$.

The $\Upsilon$ robot generates movement by pushing off its posterior leg,
which must be rigid enough to support itself as well as 
propel forward the center portion of its anterior wall 
(e.g.~run 12 in Fig.~\ref{fig:pressure}).
The robot loses kinetic energy, which is stored as elastic strain energy in the spring-like voxels between the wall's center and edge.
The most strain is present in the dorsal portion of the anterior wall.
The springs recoil, restoring kinetic energy and generating forward motion.
If the posterior leg is too soft, or the dorsal anterior wall too rigid, the $\Upsilon$ robot can suffer a large drop in performance.

Differences in geometry, however, shed no light on why (the reduced) stress-adaptive robots are more robust than nonadaptive robots:
the level of significance $(P<0.01)$ does not change after removing the only nonadaptive robot that could possibly be classified as $\Upsilon$ (Fig. \ref{fig:none}, run 19). 
Thus we continue our investigation by analyzing how stress and pressure might differentially affect the rate of developmental reactions.

\subsection*{Canalization.}

One indication of canalization \citep{waddington1942canalization,kriegman2017morphological} in our system is given by the magnitude of $\alpha_i$ in each voxel, as defined by Eqs. \ref{eq:stress} and \ref{eq:pressure}.
However, this is but one of two necessary ingredients for a developmental reaction: it indicates bodywide responsiveness to \textit{potential} stimuli, but ignores the \textit{actual} stimulus.

Thus, as proxy for canalization, we measured the amount of morphological change in reaction to local stimulus, during evaluation.
More precisely, we recorded the mean, across the body, of relative lifetime change in stiffness
\begin{equation}
\label{eq:mu}
M_{\text{body}} = \frac{1}{\#\gamma} \sum_{i \in \gamma} \left| k_i^{+}/k_i^{\circ}-1 \right| ,
\end{equation}
where 
$k_i^{\circ}$ 
is the congenital stiffness,
$k_i^{+}$ is the final stiffness, and
$\gamma = \{i : g_i = 1 \}$ contains the coordinates $i$ of each voxel $g_i$ present in the (bit array) geometry which has cardinality $\#\gamma$ (total voxels). 
Less change---lower $M_{\text{body}}$---indicates more canalization.

On average, voxels in stress-adaptive robots change their relative stiffness less than voxels in pressure-adaptive robots $(P<0.001)$ (Fig. \ref{fig:run_champs}D).
In other words, developmental reactions are canalized to a greater extent in stress-adaptive robots.
It follows, then, that the treatment with increased robustness was also the treatment with increased canalization.

To get a sense of the consistency of developmental reactions, as they occur \textit{across the body} of evolved robots, we also recorded the spatial variance of this relative lifetime change 
\begin{equation}
\label{eq:sigma}
V_{\text{body}} = \text{Var}_{i \in \gamma} \left(\,\left| k_i^{+}/k_i^{\circ}-1 \right|\,\right) .
\end{equation}
By this measure, stress-adaptive robots exhibit more uniform reactions than pressure-adaptive robots $(P<0.05)$ (Fig. \ref{fig:run_champs}E).

Taken together, then, we may say that the developmental reactions of stress-adaptive robots are more uniform in space 
(lower $V_{\text{body}}$; Fig. \ref{fig:run_champs}E), 
and more canalized in magnitude (lower $M_{\text{body}}$; 
Fig. \ref{fig:run_champs}D) than those of pressure-adaptive robots.
Pressure-adaptive robots therefore experience larger and more localized changes in stiffness during their lifetime.

There are two possibilities that could explain this more localized change in stiffness in the pressure-adaptive robots.
One possibility is that there is greater variance among the $\alpha_i$ in the pressure-adaptive robots.
The alternative is that there is greater variance in the application of pressure throughout the body.
To test the first possibility, we first normalized $\alpha_i$ in pressure- and stress-adaptive robots
by the differing ranges of $\alpha$ that evolved in the pressure-adaptive (-5.36 to 5.63) 
and stress-adaptive (-10.00 to 6.42) robots.
Then we took the variance of $\alpha_i$ across the body of each run champion, individually:
\begin{equation}
\label{eq:sigma-gain}
V_{\text{gain}} = \text{Var}_{i \in \gamma} \tilde{\alpha_i} \,, \quad \text{where} \;\, \tilde{\alpha_i}=\frac{\alpha_i-\alpha_{\text{min}}}{\alpha_{\text{max}}-\alpha_{\text{min}}}  \,.
\end{equation}
We found no evidence to support the hypothesis that $\alpha_i$ in pressure-adaptive robots vary more (or less) in space than those in stress-adaptive robots (Fig. \ref{fig:run_champs}F).

Therefore, because there is no difference in the variation of $\alpha_i$ (Fig. \ref{fig:run_champs}F), and because  $\alpha_i$ cannot change during operation (Eqs. \ref{eq:stress} and \ref{eq:pressure}), it follows that pressure was generally much more localized within
the bodies of the pressure-adaptive robots than stress within the bodies of the stress-adaptive robots.
In other words, the entire body plan encountered stress, but only a small portion of the body encountered appreciable pressure.
(An example of this localized response to localized pressure can be seen in Fig. \ref{fig:calluses}.)
We hypothesize that this global spread of stress is the likely cause of increased robustness in the stress-adaptive robots
(Fig. \ref{fig:run_champs}C).



%% file: discussion.tex
\section{Discussion}
\label{sec:discussion}

Building systems that are robust in the face of changing environmental conditions is a grand challenge in robotics and AI.
The brittleness of current systems is exemplified by the growing literature on adversarial examples
\citep{szegedy2013intriguing,
nguyen2015deep,
athalye2017synthesizing}, 
and the fact that almost all practical robots are confined to the perfectly flat floors they clean, or the hermetic factories built around their work.
Robustness is not unknown in human-engineered systems, but it is relatively rare; in nature it is everywhere, and one of the reasons is that in nature organisms develop: 
They constantly change not just their cognitive architectures but the morphologies that contain them and mediate with the external world.

It has been shown for rigid robots \citep{bongard2011morphological}
that morphological development can in some cases increase robustness since it exposes evolution to 
richer sensory information: the robot must maintain locomotion while changing its body. 
Soft robots have much greater potential in this domain:
If soft, there are more ways that morphology can change, so by definition the increase
in breadth in sensorimotor experiment induced by development will be even greater than that
for developing yet rigid machines.
Toward this goal, by allowing material stiffness to be plastic, we have here 
investigated a heretofore unexplored dimension of morphological change (stiffness) not available
to rigid robots.

Advances in materials science and 3D printing promise new engineered systems{\textemdash}protean machines{\textemdash}that may continuously morph in response to changing environmental signals.
Simply put, if a robot always changes its strategy along many morphological and neural modalities, 
it is more difficult to fool with a static adversarial example or a new task environment.
Little to no analysis has been conducted, however, into how such systems should respond to environmental stimuli in order to adapt their functions in the face of changing environmental conditions.

In initiating such a study here, we have shown that it is not just a matter of reacting to \textit{any} stimulus: different types of developmental feedback loops elicit different \textit{evolved} properties.
We observed that if one modality (stiffness) responds to one particular internal signal (engineering stress) but not another (pressure), robots evolved structure that  intrinsically buffered large deviations from their expected material properties.

Pressure and stress bear distinct mechanical load signatures which in turn stimulated very different developmental reactions.
Intriguingly, increased robustness was correlated with increased canalization: developmental reactions with stress were canalized to a greater degree than those with pressure.
Although developmental reactions with pressure did not afford the evolution of robustness here, it did increase evolutionary divergence: pressure-adaptive robots evolved more diverse 
(congenital) 
shapes than stress-adaptive robots.
Our work here suggests
there may be other developmental feedback loops that could be made available to evolution
that would lead to more diverse and robust robots.

For our purposes,  `morphology' is a robot body,
but the concepts here could equally be applied to non-embodied systems, such as the architectures of deep 
neural networks \citep{miikkulainen2017evolving,zoph2016neural}.
One could define 
internal neural processes such as node sharpening \cite{french1994dynamically},
Hebbian learning,
or neurotransmitter diffusion \cite{husbands1998better, velez2017diffusion}
as interoceptive signals to which
the neural network developmentally responds in a structural manner,
such as adding or removing neurons.
Meanwhile, at a faster time scale, synaptic weights might be tuned in response to 
exteroceptive signals such as gradients of a loss function.
Finally, such a network could be placed inside a robot
which itself is experiencing morphological change.

%% file: acknowledgments.tex
\section{Acknowledgments}

This work was supported by 
Army Research Office award W911NF-16-1-10304
and DARPA contract HR0011-18-2-0022. 
The computational resources provided by the 
UVM's 
Vermont Advanced Computing Core 
(VACC)
are gratefully acknowledged.

